\documentclass{article}
\usepackage{spconf,amsmath,graphicx,booktabs,multirow,xcolor,tikz}
\usepackage[hidelinks]{hyperref}
\usepackage{enumitem}
\usetikzlibrary{positioning,arrows.meta,fit,backgrounds}

\newcommand{\sysname}{DuplexJev}
\newcommand{\MIXCEG}{89.5}\newcommand{\MIXCEZ}{81 / 76}\newcommand{\MIXKDG}{89.9}\newcommand{\MIXKDZ}{90 / 86}\newcommand{\MIXKDE}{90.0}\newcommand{\MIXKDQ}{$-$1}\newcommand{\MIXKDM}{$-$16}

\title{Batched Speech Decisions Without Decoding:\\
Single-Token Supervision Lets a Frozen LLM Hear Beyond the Transcript}

\name{Jie Jin$^{1,\star}$, Ziyin Ma$^{2}$, Min Yin$^{3}$, Jinyu Chen$^{4}$, Haigang Song$^{1}$, Zhikun Pang$^{1}$, Xiaowen Zhang$^{1}$}
\address{$^{1}$Adventists.ai, Melbourne, Australia\\
$^{2}$College of Computer Science and Technology, Zhejiang University, Hangzhou, China\\
$^{3}$Future Design Lab, Innovation Center of Yangtze River Delta, Zhejiang University, Jiaxing, China\\
$^{4}$University of Virginia, Charlottesville, VA, USA\\
{\small $^{\star}$Corresponding author: jiejin@adventists.ai}}

\begin{document}
\ninept
\maketitle

\begin{abstract}
Full-duplex voice agents make many small, closed decisions, which current systems answer by slow autoregressive decoding. We propose \sysname{}, which feeds ASR-encoder hidden states through a small connector into a frozen LLM and reads each question as a single-token distribution over its options. Nothing is decoded, and an 8-GPU node answers 80 decisions about eight utterances in about 0.1\,s. With a last-layer connector, spoken QA stays close to reading the transcript (90\% vs.\ 91\%). \sysname{} also hears the speaker: gender and emotion accuracy both reach 90\% (from 55\% and 28\%) with a cross-attention connector, whose spoken QA drops by only 1 point (83\% to 82\%). We train decisions with cross-entropy on the read-out answer token, instead of the usual transcript distillation, whose teacher never hears the voice, and keep distillation for content. Encoders and LLMs are interchangeable; we release weights, training recipe, a batched-inference pipeline for full-duplex serving and a bilingual spoken-QA set.
\end{abstract}

\begin{keywords}
full-duplex dialogue, speech LLM, speech-to-decision, decision models, paralinguistics
\end{keywords}

\section{Introduction}
\label{sec:intro}

A full-duplex voice agent listens while it speaks. Around every user utterance it must answer small, closed questions: \emph{Is the turn complete? Is this a backchannel or a barge-in? Which pre-recorded filler fits? Who is speaking?} Most of this work is not \emph{speech-to-text} but \emph{speech-to-decision}: the agent rarely needs the transcript, only the answer to a question about it. The cost is real: in a deployed in-car assistant (Sec.~\ref{sec:setup}), such decisions are made by generating text and consume 30\% of all LLM input tokens. These decisions block the reply: the agent cannot speak until they are answered, so their latency adds to every turn. Keeping them fast under peak load requires spare capacity, which makes them a major share of both latency and cost.

Current systems pay for these decisions with autoregressive decoding. End-to-end full-duplex models~\cite{dgslm,moshi,syncllm} decode on every frame of every call, so each call holds its own decoding stream and batched serving is expensive; their decisions are also not exposed as typed outputs. Cascades transcribe and then generate an answer, which is slow and discards everything in the voice. Dedicated turn-taking classifiers~\cite{easyturn} are cheap but answer one fixed question each.

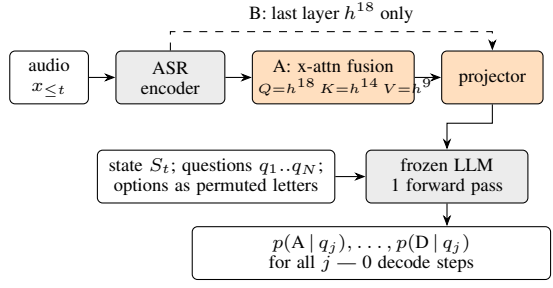
\begin{figure}[t]
\centering
\begin{tikzpicture}[
  font=\scriptsize, node distance=2.5mm and 3.5mm,
  blk/.style={draw, rounded corners=2pt, align=center, minimum height=7mm, inner sep=2pt},
  frz/.style={blk, fill=gray!15},
  trn/.style={blk, fill=orange!25},
  arr/.style={-{Stealth[length=1.5mm]}}]
\node[blk, text width=9mm] (wav) {audio\\$x_{\le t}$};
\node[frz, right=of wav, text width=13mm] (enc) {ASR\\encoder};
\node[trn, right=of enc, text width=20mm] (fus) {A: x-attn fusion\\{\tiny $Q{=}h^{18}\,K{=}h^{14}\,V{=}h^{9}$}};
\node[trn, right=of fus, text width=12mm] (proj) {projector};
\node[frz, below=6mm of proj, text width=20mm, xshift=-6mm] (llm) {frozen LLM\\1 forward pass};
\node[blk, left=4mm of llm, text width=30mm] (sch) {state $S_t$; questions $q_1..q_N$;\\options as permuted letters};
\node[blk, below=3mm of llm, text width=46mm, xshift=-10mm] (out) {$p(\text{A}\,|\,q_j),\dots,p(\text{D}\,|\,q_j)$ for all $j$ --- 0 decode steps};
\draw[arr] (wav) -- (enc);
\draw[arr] (enc) -- (fus);
\draw[arr] (fus) -- (proj);
\draw[arr] (proj.south) -- ++(0,-2.5mm) -| (llm.north);
\draw[arr] (sch) -- (llm);
\draw[arr] (llm) -- (llm|-out.north);
\draw[arr, dashed] (enc.north) -- ++(0,2.5mm) -| node[pos=0.25,above]{B: last layer $h^{18}$ only} (proj.north);
\end{tikzpicture}
\caption{\sysname{}. Only the connector (fusion block and projector, orange) is trained. Variant A fuses three encoder layers by cross-attention; variant B projects the final layer only. Every question is read as a closed-set next-token distribution. Content is aligned by transcript distillation; decisions are trained with answer-token cross-entropy (Sec.~2.1).}
\label{fig:arch}
\end{figure}

We ask whether a frozen, general-purpose LLM can answer such questions directly from speech, in the spirit of the ``System-One'' decision models~\cite{jev,openjev} that motivated this work: given a state and a question declared at run time, return a distribution over its options without generating text. If speech enters the LLM as continuous embeddings, no ASR decoder is needed. If each question lists its options under letters, the answer is the next-token distribution over those letters, so nothing is decoded. Such one-step questions batch across questions and calls, and questions that share a long dialogue context can be packed into one row (Fig.~\ref{fig:arch}).

Reading decisions from audio exposes a second problem. Connectors that feed speech into a frozen LLM are usually trained by distillation from the LLM's own output on the transcript~\cite{ultravox,diva}. This teacher never hears the voice, so the student cannot learn what is not in the words: after 6k steps, speaker gender is still at chance, although a linear probe reads it at 97\% from the encoder (Fig.~\ref{fig:r3}). Supervising only the answer token that is read out, while keeping distillation for content, fixes this.

\textbf{Contributions.}
\begin{enumerate}[leftmargin=*,nosep,label=(\arabic*)]
\item \textbf{Decoding-free batched decisions.} One forward pass, no decode steps, and exact prefix sharing for long contexts; an 8-GPU node answers 80 decisions about eight utterances in about 0.1\,s.
\item \textbf{Why speech LLMs cannot hear the speaker, and a fix.} Answer-token supervision teaches a frozen LLM gender and emotion through two connector designs; controlled comparisons of objectives and layer-wise probes locate the bottleneck in the objective, not the connector.
\item \textbf{Modularity and release.} Any ASR encoder and any frozen LLM are joined by a small connector. We release the weights, training recipe, a batched-inference pipeline for full-duplex serving and the qa100 test set.\footnote{\url{https://github.com/adventists-ai/duplexjev}; \texttt{pip install duplexjev}.}
\end{enumerate}

\section{Method}
\label{sec:method}

\subsection{Connector}
A frozen ASR encoder produces layer-wise hidden states $h^{\ell}_{1:T}$. We compare two ways of turning them into LLM input embeddings (Fig.~\ref{fig:arch}). \textbf{B (last layer)} feeds $h^{L}$ to an Ultravox-style projector~\cite{ultravox} (frame stacking + MLP) into the LLM embedding space ($d=5120$). \textbf{A (cross-attention fusion)} adds a fusion block before the projector: queries come from the final layer ($\ell=18$ for Qwen3-ASR-0.6B~\cite{qwen3asr}), keys from layer 14 and values from layer 9. The output projection is zero-initialised and added residually to $h^{L}$, so A starts identical to B. A tests the common intuition that ASR fine-tuning pushes the final layer towards linguistic content while intermediate layers keep speaker and prosodic cues~\cite{salmonn}; for this encoder our probes do not support it (Sec.~\ref{sec:para}). Both LLM and encoder stay frozen; only fusion and projector are trained.

\textbf{Training.} For content we align the connector by transcript distillation~\cite{hinton,ultravox} (R1--R2): given the audio embeddings, the frozen LLM should match the distribution it produces when the audio is replaced by the gold transcript (token-level KL, temperature 2). This keeps the LLM's behaviour on content intact, but the teacher never hears the voice: for any question whose answer is not in the words (speaker, emotion), its target is the text-only prior, and the student is trained to be exactly as uninformed. DiVA~\cite{diva} aligns a connector to a frozen LLM with the same kind of text-conditioned distillation; DeSTA2~\cite{desta2} writes paralinguistic metadata into generated text descriptions used as targets; speech LLMs trained on labelled tasks~\cite{qwenaudio,salmonn} learn such cues with cross-entropy on generated answers. We differ in supervising only the single token the readout uses, which needs neither descriptions nor decoding, in showing with probes that the objective and not the connector blocks the cue, and in a mixed objective that keeps distillation for content. For decisions we therefore introduce two ingredients. \emph{Answer-token supervision}: cross-entropy on the single answer letter, at the position and under the template that the readout of Eq.~(\ref{eq:readout}) uses, with the fixed chat-template tokens before it (e.g.\ an empty reasoning block) excluded, so the letter is the only supervised token. \emph{Mixed objective}: each sample keeps its own loss; content samples (transcription, continuation) are distilled, decision samples use answer-token cross-entropy, and the two averages are added.

\subsection{Typed single-token readout}
A request holds a state $S_t$ (the audio up to time $t$, plus optional agent state) and $N$ typed questions. Each \emph{choice} question with $K$ options is rendered as a prompt that lists the options under letters A, B, \dots, with the letter order randomly permuted to remove position bias. \emph{Boolean} questions use N/Y and \emph{score} questions use 1..$K$. We run a single forward pass, take the logits at the answer position, restrict them to the $K$ option tokens and renormalise:
\begin{equation}
p(o_k \mid S_t, q_j) = \operatorname{softmax}_{k}\big(z_{\pi_j(k)}\big),
\label{eq:readout}
\end{equation}
where $\pi_j$ maps options to letters. Nothing is sampled: the typed output is always valid, and $\max_k p$ gives a confidence score that can be used to abstain.

\subsection{Batching and cost}
Because each question needs one step, $N$ questions over the same state, or over different calls, can be read together. The simplest implementation copies the whole prompt into every row (\emph{batch-repeat}), so each row re-encodes the audio and any dialogue context. With \emph{prefix sharing}, the longest common prefix of the $N$ prompts (template, dialogue context, audio) of length $P$ is encoded once into a KV cache. The $N$ question suffixes are then concatenated into a single row; a block-diagonal 4-D mask lets each suffix attend to the prefix and causally to itself, but not to other suffixes, and position ids restart at $P$ for every suffix. The answer to question $j$ is read at the last token of its suffix. The pass costs $\approx C(P) + \sum_j C(L_j)$ instead of $N\,C(P + \bar L)$, and the KV cache holds $P + \sum_j L_j$ positions rather than $N(P+\bar L)$, so long shared contexts do not multiply memory. The construction is exact up to numerical precision (Sec.~\ref{sec:results}).

By contrast, a full-duplex model spends a decoding step on every frame of every call whether or not a decision is due, and a cascade spends $O(\text{transcript} + \text{answer})$ autoregressive steps per decision.

\subsection{Deployment pattern}
\label{sec:deploy}
Because a pass is one step with no decoding, \sysname{} could also run on a fixed clock rather than per detected utterance: every tick (e.g.\ 160--200\,ms) the server collects the newest audio of all connected calls and answers their decision sets in one batched pass, with turn-taking (turn complete, backchannel, barge-in) as one of the questions, so that it could replace a separate VAD or end-pointer; we do not evaluate streaming input here. Business decisions (filler, routing, history) share the pass. Fillers show the gain: instead of generating one (the deployed assistant spends a 5k-token LLM call on a 20-token filler), a fitting filler is \emph{selected} from a large pre-recorded library, by boolean rows over a shortlist or a coarse-to-fine tree of lettered questions that grows without retraining, at the cost of one readout (Table~\ref{tab:cost}).

\section{Experimental Setup}
\label{sec:setup}

\textbf{Models.} The LLM is Qwen3-32B~\cite{qwen3}, frozen, in bf16. The encoder is Qwen3-ASR-0.6B, frozen. Trainable parameters are 21.1\,M for A (fusion 3.2\,M, projector 17.8\,M) and 17.8\,M for B. A and B are trained from scratch with identical data and schedule (A on 8$\times$H200, B on 8$\times$A800; global batch 16, lr $2{\times}10^{-4}$, 1k warm-up, cosine decay, 32k steps per round). \textbf{R1--R2} use only content tasks, transcription and continuation, with the distillation objective; the data are a public ASR/translation mixture of WenetSpeech~\cite{wenetspeech}, GigaSpeech, LibriSpeech~\cite{librispeech}, Common Voice~\cite{commonvoice} and CoVoST, split into 100 disjoint packs of which R1 and R2 each use one (0.5\,M utterances per round). \textbf{R3} is a short paralinguistic stage that starts from the R2 connector (batch 16, lr $10^{-4}$). The \emph{gender pack} has 32k real utterances from AISHELL-1~\cite{aishell} and LibriSpeech with corpus-provided labels; the \emph{emotion pack} has 26k utterances with four classes (neutral, happy, angry, sad) from ESD~\cite{esd} (Chinese and English) and CREMA-D~\cite{cremad} (English). Every clip serves three tasks in equal shares: transcription, continuation, and one closed-set question with permuted option letters and paraphrased stems. Test speakers are disjoint from all training speakers and no synthetic voices are used. With everything else fixed we compare distillation (6k steps) with answer-token supervision for gender (6k) and emotion (3k), and a four-task mixture (a content pack plus both decision tasks, 4k steps) under all-cross-entropy (MIX-CE) or the mixed objective (MIX-KD). To test an encoder swap we plug in the released Whisper-large-v3-turbo encoder and projector~\cite{whisper,ultravox}, trained for the same Qwen3-32B, without modification.

\textbf{qa100 (released).} 100 four-way multiple-choice questions: 50 Chinese and 50 English, each language split into 25 logic and 25 factual items, with the gold letter balanced over A--D. Question stems are synthesised with VoxCPM2~\cite{voxcpm}; options are given as text. Every clip is verified by an independent ASR (exact normalised match, re-synthesised otherwise). Three conditions share the same prompt: \emph{audio}; \emph{text}, where the oracle transcript replaces the audio (an upper bound for hearing); and \emph{options-only} (a lower bound, 0.36).

\textbf{Content understanding.} Besides qa100 we use the main-language part of the ZJU audio benchmark v2.0.0~\cite{zjugender} (ZJU-ML): 100 four-way questions, 50 Chinese and 50 English, whose 50 factual items are real recordings (ODSQA, SD-QA) and 50 math and logic items TTS. We also use Easy-Turn~\cite{easyturn} for zero-shot turn-taking.

\textbf{Paralinguistics.} \emph{Gender}: our 800-utterance real-speech set (AISHELL-1, LibriSpeech, Common Voice; balanced by gender and language) and the independently constructed ZJU audio-gender-benchmark v1.2.0~\cite{zjugender} described next. \emph{Emotion}: 800 utterances balanced over the four classes from held-out ESD speakers (two per language) and CREMA-D actors. Transcript-only controls are at chance by construction, since both emotion corpora read the same sentences in every emotion.

\textbf{Gender (ZJU).} The gender part of the same benchmark (v1.2.0)~\cite{zjugender}: 100 clips, 50 TTS and 50 real speech (MagicData-RAMC Chinese; SLURP and AMI English), balanced by gender, scored with the official script. We report accuracy on all 100 clips and on the real-speech subset, together with a \emph{text-only} control, which is at chance by construction and confirms the decision is made from the voice.

\textbf{Production workload.} Dimi, an in-car voice assistant operated by DoumaoAI (Qwen3-ASR-0.6B $\rightarrow$ 72B LLM $\rightarrow$ TTS; 1.45\,M user turns from 3{,}721 devices over 91 days), issues up to three LLM calls per turn only to make small decisions: a tool-need check (median 1{,}471 input tokens), an intent label (530) and a 20-token opening filler (5{,}020). The median time from end of speech to first audio is 1.29\,s, and 1.9\% of utterances already call a separate speaker-identification model.

\textbf{Cost.} All timings use one H200 and bf16. \sysname{} latency is the median of three timed passes after one warm-up in plain PyTorch (eager attention, no CUDA graphs), excluding audio loading. The \emph{cascade} baseline answers ten questions per event (e.g.\ filler among 8, route among 6, urgency 1--5, turn complete): Qwen3-ASR-0.6B transcribes greedily, then Qwen3-32B served with vLLM~\cite{vllm} (prefix caching, greedy) generates a 10-field JSON; all 30 outputs parse. Events are 30 real utterances, 10 each of 2--3, 4--6 and 8--12\,s, from the speaker-disjoint gender test set. To separate the readout from the implementation, we also run the same ten questions on the same vLLM engine as single-token readouts (one lettered prompt per question, one output token, answer from the logprobs), with the transcript in place of the audio embeddings. Our connectors feed 6.25 audio tokens/s (stack factor 2), 10--33 more tokens per row than the transcript here (rows of 47--67 tokens), so this slightly underestimates the cost of \sysname{} on an optimised engine, before the saved ASR; the PyTorch timings and prefix-sharing runs use earlier connectors with stack factor 8 (${\approx}1.6$ tokens/s). We repeat both with a 1{,}471-token dialogue context, the median input of the production tool-need check.

\section{Results}
\label{sec:results}

\subsection{Batched readout and cost}

\begin{table}[t]
\centering
\caption{Ten decisions per event, one H200, Qwen3-32B on vLLM, 30 real utterances. \emph{JSON}: one JSON answer generated from the ASR transcript (ASR, 411\,ms, not included). \emph{Readout}: one single-token prompt per question on the transcript (approximates \sysname{}). Capacity: events/s whose batch finishes within the budget.}
\label{tab:cost}
\setlength{\tabcolsep}{3pt}
\begin{tabular}{llrrrrr}
\toprule
 & & decode & 1 event & \multicolumn{3}{c}{events/s within} \\
\cmidrule(lr){5-7}
ctx & method & steps & (ms) & 0.25\,s & 0.5\,s & 2\,s \\
\midrule
0 & JSON & 86 & 1567 & 0 & 0 & 16.9 \\
0 & readout & 0 & 92 & 17.3 & 20.4 & 33.5 \\
1.5k & JSON & 87 & 1598 & 0 & 0 & 8.4 \\
1.5k & readout & 0 & 144 & 9.7 & 12.0 & 18.4 \\
\bottomrule
\end{tabular}
\end{table}

\textbf{Latency and capacity.} On the same engine, reading ten decisions as single tokens takes 92\,ms per event, against 1.57\,s to generate them as JSON, a 17$\times$ reduction before counting the 411\,ms of autoregressive transcription the cascade also needs (Table~\ref{tab:cost}). With a 1{,}471-token dialogue context, the median of the production tool-need check, the gap is 11$\times$. The JSON cascade cannot serve any load within 0.5\,s, whereas the readout serves 17--20 events/s (10--12 with context) on one GPU. Replicated on one 8$\times$H200 node (one engine per GPU), eight events (80 decisions) finish in 94--103\,ms, and the readout serves 161 events/s, i.e.\ 1{,}613 decisions/s, within 0.5\,s (100 events/s with the 1.5k context) and saturates at 3{,}684 decisions/s. At full saturation (64 simultaneous events) the two are on par (335 vs.\ 311 decisions/s without context), since batched decoding is cheap and one JSON prompt amortises the dialogue context over all ten questions, while our readout repeats the transcript and instructions in every question row. The advantage is latency and capacity under a conversational budget, not raw throughput.

\textbf{Prefix sharing is exact and removes the context cost.} Shared and one-by-one readouts give the same answer on 195 (A) and 198 (B) of 200 question--utterance pairs (20 utterances $\times$ 10 questions), and on 40 of 40 with 0.5--5k-token contexts. All differences but one (0.31) are near-ties (margin $\le 0.06$; median $|\Delta p|$ 0.01--0.02, bf16 noise). Without context the audio prefix is short (stack factor 8: ${\approx}1.6$ audio tokens/s, 36--66 prefix tokens), so the question text dominates and sharing halves the cost of 100 questions over one 2.6\,s utterance (3.1\,s batch-repeat vs.\ 1.6\,s; 7.6\,s sequential). With a production-sized context, batch-repeat re-encodes the context in every row (158k tokens for 100 questions over 1.5k) and runs out of memory already at 5 questions over a 5k context, while the packed row keeps the KV at $P+\sum_j L_j$ positions: 100 questions over a 1.5k context take 5.1\,s instead of 37.9\,s sequentially, and 50 over a 5k context 5.5\,s instead of 112\,s.

\textbf{Batching does not change answers.} Across different utterances in one batch, batched and one-by-one answers agree on 790 of 800 real-speech gender items and on 99--100 of 100 ZJU items, and qa100 scores move by at most one item.

\subsection{Content understanding}

\begin{table}[t]
\centering
\caption{Content-understanding accuracy (\%). qa100: released spoken QA (/100, zh/en /50 each). ZJU-ML: spoken questions from~\cite{zjugender}, half real, half TTS. Easy-Turn: zero-shot 4-way turn state. Our connectors are shown after R2 (two 1\% packs).}
\label{tab:main}
\setlength{\tabcolsep}{3pt}
\resizebox{\columnwidth}{!}{%
\begin{tabular}{lcccc}
\toprule
input & qa100 & zh / en & ZJU-ML & Easy-Turn \\
\midrule
oracle transcript & 91 & 45 / 46 & 89 & 80.5 \\
options only & 36 & 18 / 18 & 39 & -- \\
Whisper, released (Qwen3-32B) & 89 & 43 / 46 & 85 & 76.1 \\
Whisper, released (Gemma-3-27B) & 92 & -- & 77 & -- \\
\midrule
B last layer, R2 & 90 & 44 / 46 & 79 & 76.1 \\
A x-attn, R2 & 83 & 39 / 44 & 77 & 77.1 \\
\bottomrule
\end{tabular}}
\end{table}

\textbf{The interface does not cost understanding.} With the last-layer connector (B), after only two 1\% packs, \sysname{} answers 90 of 100 spoken questions against 91 when reading the oracle transcript (Table~\ref{tab:main}); English is identical, and the text-condition errors are arithmetic or date items a single token without reasoning misses. A reaches 83; both reach 77--79 on ZJU-ML, below the released Whisper connector (85), trained on far more data. Swapping parts leaves the readout unchanged: the Whisper connector gives 89, and its Gemma-3-27B version 92 (transcript bound 93). Zero-shot turn-taking reaches 76--77\% on Easy-Turn~\cite{easyturn} from audio against 80.5\% from text; recall on \emph{wait} stays near 0.3 even from text, a limit of the LLM's reading rather than of hearing. Confidence gates abstention: answering only at $p\ge0.95$ covers 87\% of qa100 at 93\% accuracy (B).

\subsection{Paralinguistics: distillation vs.\ answer supervision}
\label{sec:para}

\begin{table}[t]
\centering
\caption{Paralinguistic accuracy (\%) and change in content understanding. Gender: our 800 real utterances / ZJU native / ZJU lettered. Emotion: 4-way, 800 utterances. $\Delta$: change vs.\ the R2 start of the same connector. Probes: logistic regression on pooled features.}
\label{tab:para}
\setlength{\tabcolsep}{2pt}
\resizebox{\columnwidth}{!}{%
\begin{tabular}{lccccc}
\toprule
 & gender & ZJU n/l & emotion & $\Delta$qa100 & $\Delta$ZJU-ML \\
\midrule
probe, encoder last layer & 97.5 & -- & 92.5 & & \\
probe, connector A / B & 98.1 / 96.5 & -- & 87.9 / 88.3 & & \\
Whisper, released connector & 55.4 & 51 & 27.1 & & \\
A after R2 & 55 & 53 / 50 & 28 & & \\
B after R2 & 54 & 50 / 51 & 27 & & \\
\midrule
A, distillation, 6k & 53 & 49 / 50 & -- & & \\
B, distillation, 6k & 53 & 52 / 52 & -- & & \\
A, answer CE, gender, 6k & 89.4 & 73 / 67 & 27 & +3 & +2 \\
B, answer CE, gender, 6k & 87.9 & 60 / 57 & 26$^\ast$ & $-$3 & +2 \\
A, answer CE, emotion, 3k & 49$^\dagger$ & 40 / 43 & 71.8 & +1 & $-$6 \\
B, answer CE, emotion, 3k & -- & -- & 85.5 & $-$1 & $-$10 \\
\midrule
A, MIX-CE, 4k & \MIXCEG{} & \MIXCEZ{} & 88.9 & $-$4 & $-$17 \\
A, MIX-KD, 4k & \MIXKDG{} & \MIXKDZ{} & \MIXKDE{} & \MIXKDQ{} & \MIXKDM{} \\
\bottomrule
\end{tabular}}
\\[1pt]{\scriptsize $^\ast$B with the 80\%-gender variant. $^\dagger$below chance: emotion training alone biases the gender answer.}
\end{table}

\textbf{Distillation cannot teach what the transcript lacks.} Starting from R2, both connectors were trained for 6k steps on the gender pack with the distillation objective. Gender stays at chance (53\%) on real speech and on ZJU, and the answers keep the text prior (female recall 0.3, male 0.8). The failure is not a matter of data, capacity or schedule: a gender-only mix, a 5$\times$ higher learning rate, a re-start from a random connector, and evaluation on the \emph{training} utterances with the training prompt (56\%) all leave it at chance, while the loss keeps falling as the student matches its uninformed teacher.

\begin{figure}[t]
\centering
\includegraphics[width=\columnwidth]{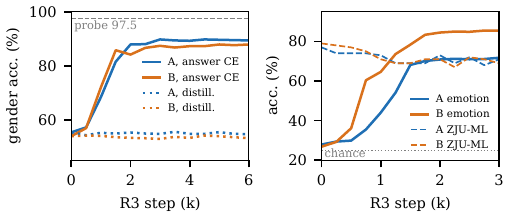}
\caption{R3 from the R2 connector, same data and schedule. Left: gender on 800 real utterances under distillation (flat) and answer-token supervision, for A and B. Right: emotion under answer-token supervision, with ZJU-ML (dashed).}
\label{fig:r3}
\end{figure}

\textbf{Answer-token supervision teaches it.} Changing only the loss, the same data and schedule take gender to 89\% (A) and 88\% (B) and emotion to 72\% (A) and 86\% (B) (Table~\ref{tab:para}, Fig.~\ref{fig:r3}). With 80\% gender questions the cue is learned within 250 steps (A: 92\%, ZJU native 97\%). Gender training moves content scores by $-$3 to +3 points. Cross-task checks show that the connector learns the cue, not an answer bias: gender-trained models stay at chance on emotion (26--29\%), and the emotion-trained model does not know gender.

\textbf{The connector is not the bottleneck.} Probes find gender (97--99\%) and emotion (88--93\%) already linearly decodable in the encoder's last layer and in the embeddings both connectors hand to the LLM. Consistently, A and B learn gender at the same rate (2k steps: 88 vs.\ 84\%), and on emotion B is even better: the objective decides.

\textbf{Keeping content understanding.} On qa100 all answer-supervised runs stay within 3 points of their start (all-cross-entropy mixing: $-$4; Table~\ref{tab:para}). ZJU-ML is more sensitive. Gender training alone is neutral on it (+2), whereas emotion data cost 6--10 points and both four-task mixtures 16--17, almost all on its real-speech factual half (e.g.\ MIX-KD: $-$15 of 50 there, $-$1 on the TTS half), in Chinese and English alike. The mixed objective does not reduce this drift, but it keeps qa100 ($-$1 vs.\ $-$4) and gives the best paralinguistic scores (gender 90, ZJU 90/86, emotion 90). 

\section{Conclusion}
\label{sec:conclusion}
\sysname{} reads many typed decisions from speech in one forward pass of a frozen LLM. Transcript-distilled speech LLMs are deaf to the speaker because their teacher cannot hear the voice; supervising the token that is read out lets the LLM use the gender and emotion cues the encoder carries. On-device gating with small LLMs is next.

\section{Compliance with Ethical Standards}
Public corpora and benchmarks are used under their licences (emotion corpora for research only); no new human-subject data were collected. Labels are corpus-provided; the classifiers study information flow, not individuals.

\section{Acknowledgements}
This work was funded by Adventists.ai. Claude (Anthropic) assisted with code.

\bibliographystyle{IEEEbib}
\bibliography{refs}

\end{document}